\pdfoutput=1

\documentclass[11pt]{article}

\usepackage[final]{acl}

\usepackage{times}
\usepackage{latexsym}
\usepackage{longtable}
\usepackage{placeins}

\usepackage{needspace}
\usepackage[T1]{fontenc}

\usepackage[utf8]{inputenc}

\usepackage{microtype}

\usepackage{inconsolata}

\usepackage{graphicx}

\title{iShumei-Chinchunmei at SemEval-2025 Task 4: A balanced forgetting and retention multi-task framework using effective unlearning loss}

\author{
  Yujian Sun\textsuperscript{1},
  Tian Li\textsuperscript{2}
\\
  \textsuperscript{1}Shumei AI Research Institute, Beijing, China\\
  \textsuperscript{2}School of Computing, Newcastle University, Newcastle upon Tyne, UK
\\
  \texttt{sunyujian@ishumei.com},
  \texttt{t.li56@newcastle.ac.uk}
}

\begin{document}
\maketitle
\begin{abstract}




As the Large Language Model (LLM) gains widespread adoption, increasing attention has been given to the challenge of making LLM forget non-compliant data memorized during its pre-training. Machine Unlearning focuses on efficiently erasing sensitive information from LLM under limited computational resources. To advance research in this area, SemEval 2025 Task 4: "Unlearning Sensitive Content from Large Language Models" introduces three unlearning datasets and establishes a benchmark by evaluating both forgetting effectiveness and the preservation of standard capabilities. In this work, we propose a more controllable forgetting loss, Effective Unlearning Loss, and explore its integration with various techniques to achieve more efficient and controlled unlearning. Our system ultimately ranked 5th on the competition leaderboard.

\end{abstract}

\section{Introduction}

Large Language Models (LLMs) have achieved remarkable success across various natural language tasks. However, LLMs tend to memorize sensitive information from their training data, posing potential risks such as privacy breaches and copyright violations \cite{wang2024machine}. Malicious attacker can exploit this vulnerability to extract confidential content, leading to unintended exposure. Machine Unlearning has emerged as a research field to address this issue, focusing on the following core challenges\cite{qu2023learn, li2025machine}:(1) Preserving essential information and capabilities while ensuring the removal of targeted data. (2) Adapting to different data types. (3) Balancing computational cost and efficiency.


SemEval-2025 Task 4 introduce the challenge "Unlearning Sensitive Content from Large Language Models," aiming to establish a robust benchmark for evaluating the effectiveness of unlearning strategies in LLMs \citep{ramakrishna2025lumellmunlearningmultitask,ramakrishna2025semeval2025task4unlearning}. The task encompasses three data categories: long-form synthetic creative documents with different genres, short form synthetic biographies containing personal information, and real documents sampled from the target model’s training dataset. Each dataset includes predefined "Forget" and "Retain" sets, and encompasses two evaluation tasks: sentence completion and question-answering. The evaluation not only assesses the success of unlearning but also measures the impact on general capabilities using the MMLU Benchmark. To encourage a balance between computational efficiency and performance, the organizers also impose runtime constraints on submitted solutions.


In this competition, we propose Effective Unlearning Loss (EUL). This aims to erase knowledge related to the data to be forgotten by perturbing the model's gradients during training. We integrate this technique with the standard Supervised Fine-Tuning (SFT) process into a multi-task learning paradigm to ensure controllable unlearning. Additionally, various data processing and augmentation strategies \cite{choi2024snap, shi2024ulmr} are explored to see their impact on the final performance.

\begin{figure*}[!t]
  \centering
  \includegraphics[width=2\columnwidth]{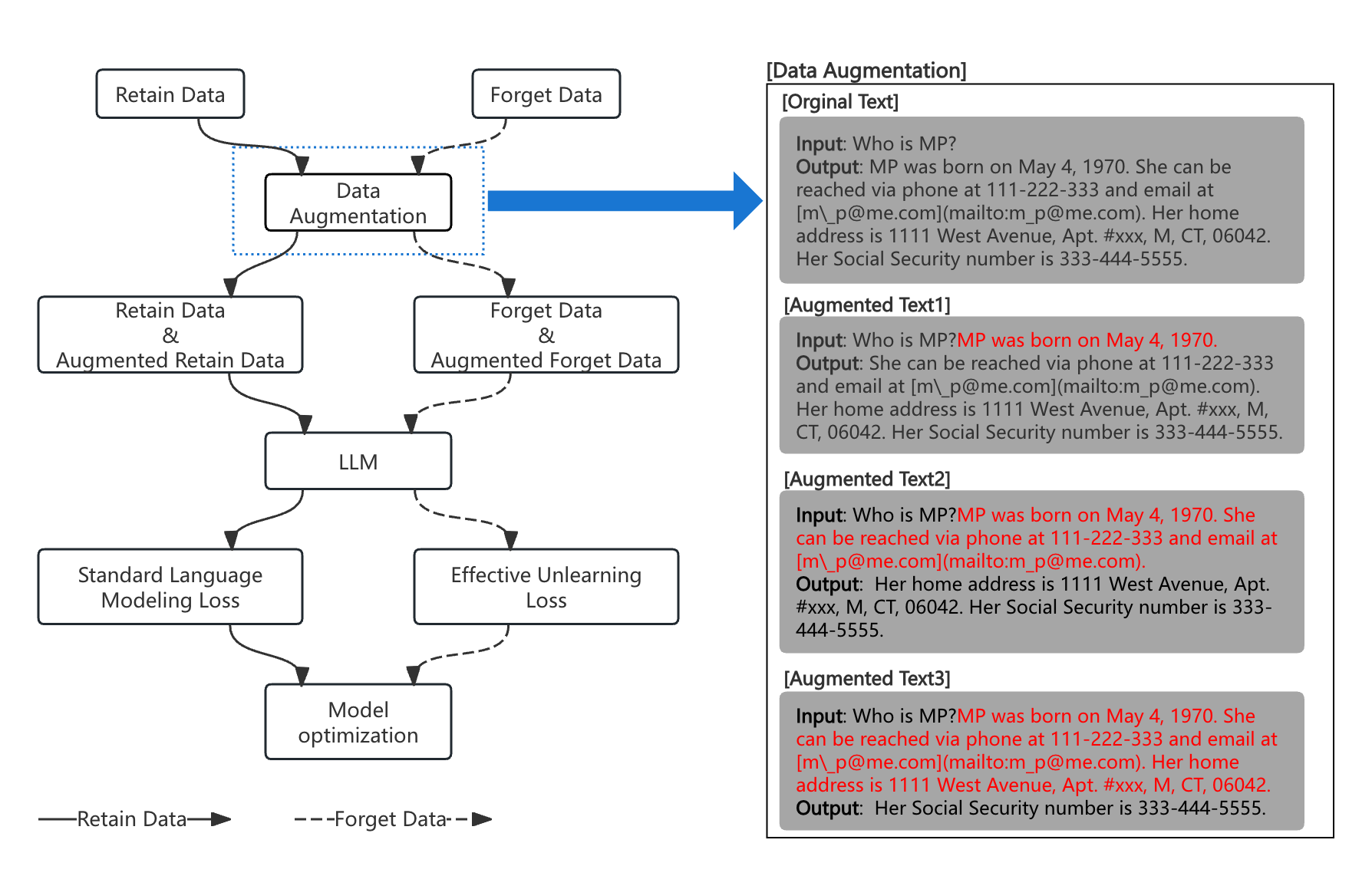} 
  \caption{The Overview of Our System.}
  \label{fig:overview}
\end{figure*}

Our contributions are as follows:
\begin{itemize}
	\item \textbf{Introduction of EUL}: This loss is the multiplicative inverse of the original SFT loss. Its inverse property makes it effective in making LLM forget the target knowledge. We integrate it into a multi-task learning paradigm, allowing the model to perform SFT on data that should be retained and EUL on data that should be erased concurrently.
	\item \textbf{Comprehensive Exploration}: We thoroughly investigate the performance of our method under different configurations and data processing/augmentation settings, providing a reliable reference for future research.
	\item \textbf{Competitive Performance}: Our approach secures 5th place on the final leaderboard, demonstrating its effectiveness.
\end{itemize}

\section{Background}

Research on knowledge unlearning in large language models (LLMs) is an emerging field, with fine-tuning-based unlearning being one of the most common method. This involves retraining the model on datasets containing specific target knowledge to weaken its memory of the unwanted information.

One intuitive unlearning method involves gradient ascent on the forget data, which increases the loss on that data to force the model to forget specified knowledge. However, this often leads to optimization instability and poor performance. To solve this, \citet{veldanda2024llm} propose a comprehensive training approach that involves gradient ascent, standard gradient descent on the forget dataset, and minimizing KL divergence to maintain the model’s performance on retained knowledge. Similarly, \citet{jang2022knowledge} introduces a gradual gradient ascent approach, which stabilizes the unlearning process and avoids instability. 

Another unlearning method involves replacing the forgotten knowledge, such as \citet{choi2024snap} and \citet{shi2024ulmr} who replace forgotten answers with negative responses like "I don't know," while \citet{eldan2023s} uses reinforcement learning to identify and replace key phrases. However, \citet{mekala2024alternate} warns that relying solely on negative feedback for unlearning may result in non-sensical outputs and introduce privacy risks, reducing the model's effectiveness.

In response, we propose a more pragmatic approach to unlearning in LLMs that combines multi-task learning, data augmentation, and EUL to facilitate faster and more efficient knowledge forgetting under constrained resources.

\section{System Overview}


In this competition, we propose the Effective Unlearning Loss (EUL), which, combined with traditional Supervised Fine-Tuning (SFT), forms a multi-task learning framework. Meanwhile, we observe a significant performance discrepancy between long and short outputs. To mitigate this issue, we incorporate data augmentation. The overall workflow is illustrated in Figure~\ref{fig:overview}.

Section~\ref{sec:EUL} provides a detailed introduction of EUL, while section~\ref{sec:MTL} demonstrates its integration with the original SFT process to enable multi-task training. Finally, section 3.3 discusses our data augmentation strategies.




\subsection{Effective Unlearning Loss}
\label{sec:EUL}


Unlike traditional gradient ascent, we redesign a novel loss function, EUL, which enables the model to achieve the forgetting effect even during gradient descent. The formulation is as follows:

\begin{equation}
L_{EUL}=\alpha  \times  \frac{1}{L_{ntp}(x_{input},y_{forget})}
\end{equation}



Here, $L_{ntp}(x_{input},y_{forget})$ represents the next-token prediction loss used in conventional SFT, and $\alpha$ is a scaling factor. When the model's output closely aligns with the distribution of the information to be forgotten, the loss increases significantly; conversely, it remains low when the output deviates from that information. Compared to gradient ascent, which can easily lead to model instability, EUL offers a more stable and effective choice, ensuring that the induced forgetting remains within a controlled and safe range.

\subsection{Multi-Task Learning}
\label{sec:MTL}



To enable the model to forget specified content while retaining the information we want to keep, we design two tasks: the Forget-Set Task and the Retain-Set Task.
\begin{itemize}
	\item The Forget-Set Task focuses on the unlearning objective, using data that needs to be forgotten and optimizing the model with $L_{EUL}$.
	\item The Retain-Set Task ensures information retention, leveraging non-forgotten data and employing the conventional next-token prediction loss $L_{ntp}$.
\end{itemize}


To prevent interference between the two tasks, each batch contains data from only one task at a time. By alternating training between these tasks, the model achieves controlled forgetting in a stable manner.

\subsection{Data Augmentation} 
\label{sec:DA}


We observe a long-tail distribution phenomenon in output length within the competition dataset. Further experiments reveal that this imbalance adversely affects model performance. For instance, if the original output is a short text, the unlearning process effectively removes the associated knowledge. However, when the original output is a long text, the unlearning procedure may fail.



To address this issue, we propose a data augmentation strategy based on re-segmentation to balance the distribution of short and long output. Specifically, we segment long output into individual sentences and incrementally move portions of the output into the input, generating shorter outputs as training samples. This process, illustrated in figure~\ref{fig:overview} right, enhances the model's adaptability to varying output lengths.

In addition, we also try the negative response replace scheme mentioned in \cite{tofu2024}. It achieves the goal of forgetting by replacing sensitive information with the safety terms (for example, "I don't know") and tuning models on it.

\begin{table*}[!t]
  \centering
    \begin{tabular}{llll|llllll}
    \hline
    \multicolumn{4}{c}{\textbf{Model Component}} & \multicolumn{4}{c}{\textbf{Score}} \\ \hline
         RD & NR & DA & EUL & MIA & TAS & MMLU & Final \\ \hline
         × & O & × & × & 0.000 & 0.092 & 0.281 & 0.124 \\ \hline
         × & O & O & × & 0.000 & 0.092 & 0.278 & 0.124 \\ \hline
         O & O & × & × & 0.000 & 0.124 & 0.283 & 0.135 \\ \hline
         O & O & O & × & 0.000 & 0.137 & 0.278 & 0.138 \\ \hline
         × & × & × & O & 0.993 & 0.408 & 0.229 & 0.543 \\ \hline
         × & × & O & O & 0.989 & 0.421 & 0.229 & 0.547 \\ \hline
         O & × & × & O & 0.009 & 0.185 & 0.278 & 0.157 \\ \hline
         O & O & × & O & 0.000 & 0.095 & 0.285 & 0.127 \\ \hline
         O & × & O & O & 0.593 & 0.395 & 0.275 & 0.421 \\ \hline
         O & O & O & O & 0.035 & 0.222 & 0.279 & 0.179 \\ \hline
    \end{tabular}
  \caption{
    The main results in our experiments.
  }
  \label{tab:AS2}
\end{table*}

\begin{table*}
  \centering
    \begin{tabular}{|l|l|l|l|l|l|l|l|l|l|}
    \hline
        EP & LR & RD & NR & DA & EUL & MIA & TAS & MMLU & Final \\ \hline
        3 & 1.00E-04 & O & × & O & O & 0.135 & 0.245 & 0.272 & 0.217 \\ \hline
        3 & 1.00E-05 & O & × & O & O & 0.000 & 0.092 & 0.280 & 0.124 \\ \hline
        3 & 1.00E-06 & O & × & O & O & 0.000 & 0.092 & 0.275 & 0.122 \\ \hline
        4 & 1.00E-04 & O & × & O & O & 0.215 & 0.278 & 0.270 & 0.254 \\ \hline
        4 & 1.00E-05 & O & × & O & O & 0.000 & 0.112 & 0.280 & 0.131 \\ \hline
        4 & 1.00E-06 & O & × & O & O & 0.000 & 0.092 & 0.276 & 0.122 \\ \hline
        5 & 1.00E-04 & O & × & O & O & 0.593 & 0.395 & 0.275 & 0.421 \\ \hline
        5 & 1.00E-05 & O & × & O & O & 0.001 & 0.112 & 0.279 & 0.131 \\ \hline
        5 & 1.00E-06 & O & × & O & O & 0.000 & 0.092 & 0.277 & 0.123 \\ \hline
    \end{tabular}
  \caption{
    Ablation study on hyper-parameters }
\label{tab:AS1}
\end{table*}


\section{Experiment}

In this competition, the organizers require unlearning to be performed on two models: OLMo-7B and OLMo-1B \cite{groeneveld2024olmo}. The original tasks include sentence completion and question-answering. To evaluate the effectiveness of unlearning, the organizers propose the following four evaluation metrics:
\begin{itemize}
	\item \textbf{Task Aggregate Score (TAS)}: This metric measures the model's performance across various tasks.
	\item \textbf{Membership Inference Attack Score (MIA)}: This evaluates the extent to which the relevant knowledge is retained or effectively forgotten.
	\item \textbf{MMLU Score}: This measures whether the model's overall linguistic capabilities degrade after the unlearning process.
	\item \textbf{Final Score}: The average of the previous three scores.
\end{itemize}


We thoroughly explore the impact of different technical combinations on the final unlearning outcome and conduct ablation studies on the modules and hyperparameters. 
Due to space limitations, all experimental results and analyses presented in the main body of this paper are based on OLMo-1B, with the results for OLMo-7B detailed in Appendix~\ref{sec:appendixA}.
All results presented are based on data publicly released by the organizers during the competition period. The results are presented in section~\ref{sec:Result} and section~\ref{sec:Ablation}.

All training is performed using LoRA \cite{hu2022lora} for model fine-tuning, with Rank, alpha, and dropout set to 8, 32, and 0.05, respectively. 
For EUL, $\alpha$ is set as 1. For the training hyperparameters, we use the following settings:$epoch=5$, $lr=1e-4$, and $batch=32$. All experiments are conducted on a single Nvidia A100 (40G) GPU, and each unlearning process is completed within one hour.

\section{Result}
\label{sec:Result}

\begin{table*}[!ht]
  \centering
    \begin{tabular}{|l|l|l|l|l|l|l|l|l|l|l|}
    \hline
        Epoch & Step & LR & RD & NR & DA & EUL & MIA & TAS & MMLU & Final \\ \hline
        3 & 484 & 1.00E-04 & O & × & O & O & 0.135 & 0.245 & 0.272 & 0.217 \\ \hline
        6.86 & 484 & 1.00E-04 & O & × & × & O & 0.020 & 0.103 & 0.289 & 0.137 \\ \hline
        4 & 646 & 1.00E-04 & O & × & O & O & 0.215 & 0.278 & 0.270 & 0.254 \\ \hline
        9.16 & 646 & 1.00E-04 & O & × & × & O & 0.075 & 0.265 & 0.287 & 0.209 \\ \hline
        5 & 807 & 1.00E-04 & O & × & O & O & 0.593 & 0.395 & 0.275 & 0.421 \\ \hline
        11.45 & 807 & 1.00E-04 & O & × & × & O & 0.175 & 0.306 & 0.286 & 0.255 \\ \hline
    \end{tabular}
  \caption{
    Data Augmentation and Training Steps }
\label{tab:DA_TS}
\end{table*}



Table~\ref{tab:AS2} illustrates the impact of different method combinations on performance. Here, EUL denotes Effective Unlearning Loss, NR indicates that the original outputs for the forgotten data are replaced with safety terms. DA is the inclusion of data augmentation during training, RD refers to supervised fine-tuning the data in the retain set,
"×" denotes the absence of the corresponding technique, while "O" indicates its application.





The best performance is achieved when using EUL to process forgotten data while incorporating data augmentation and fine-tuning retained data. This success can be attributed to the multi-task training framework, which effectively balances forgetting and retention.
Note that although the approaches employing only EUL and the combination of EUL and DA achieve relatively high scores, this performance comes at the significant compromise on the MMLU metric. 
Such a substantial degradation contradicts the evaluation criteria from the organizers, so we discard them.
We derive the following key findings:
\begin{itemize}
	\item Fine-tuning on retained data is essential, regardless of the chosen technique combination. Experiments omitting RD resulted in a significant drop in MMLU scores, indicating severe degradation in the model’s linguistic capabilities.
	\item Although adding NR improves MMLU, it negatively impacts the unlearning effectiveness. This is due to the small dataset size, making it highly prone to overfitting. In contrast, EUL achieves a better balance between MMLU and unlearning performance.
	\item Incorporating DA enhances the MIA metric, albeit with a slight trade-off in MMLU. However, since the final score is the average of the three metrics, DA provides an overall positive impact.
\end{itemize}


Due to computational constraints, the most balanced result we obtained before the competition deadline was (RD\&NR\&EUL). Therefore, our result on the leaderboard is (RD\&NR\&EUL) result, not our best result. 
Our best result was (RD\&DA\&EUL), which showed consistent performance across both OLMo-1B and OLMo-7B. The detailed results can be found in Appendix~\ref{sec:appendixA}.



\section{Ablation Study}
\label{sec:Ablation}


We also explore the impact of different hyperparamter settings, including learning rate, epoch, magnitude scaling of EUL, and the effectiveness of data augmentation.
We present the results in the following sections, which are based on OLMo-1B. The trends for OLMo-7B are consistent with those of OLMo-1B, and all results for both OLMo-1B and OLMo-7B can be found in the Appendix~\ref{sec:appendixA}.
\subsection{Hyper-parameters ablation}  

Table~\ref{tab:AS1} demonstrates the performance variations under different hyperparameters. Due to computational resource limitations, we are only able to test results up to a maximum of 5 epochs. However, it is evident that as the epoch number and learning rate increase, the performance increases

\subsection{Data Augmentation and Training Steps}

To examine whether the performance improvements from data augmentation are solely due to the increased number of training steps, we conducted a series of controlled experiments, as shown in Table~\ref{tab:DA_TS}. Specifically, we compared models trained with data augmentation for 3, 4, and 5 epochs to models trained for the same number of steps without data augmentation. Although both increasing the number of training steps and applying data augmentation improve prediction performance, the benefit of data augmentation outweighs that of simply increasing training steps.


\subsection{Amplify EUL} 
To further examine the impact of our EUL approach, we square the EUL to produce more extreme upper and lower bounds for the loss variation. However, as demonstrated in table~\ref{tab:accents}, this magnitude scaling does not yield any noticeable performance improvement.

\begin{table}\footnotesize
  \centering
    \begin{tabular}{l|l|l|l|l}
    \hline
        ~ & MIA & TAS & MMLU & Final \\ \hline
        EUL & 0.59 & 0.39 & 0.28 & 0.42 \\ \hline
        EUL$^2$ & 0.39 & 0.22 & 0.28 & 0.29 \\ \hline
    \end{tabular}
  \caption{Further exploration of EUL capabilities.}
  \label{tab:accents}
\end{table}

\section{Conclusion} 
The organizers of SemEval-2025 Task 4 introduce three Machine Unlearning datasets, design to assess unlearning capabilities from multiple perspectives and across various data types. This dataset effectively highlights the key challenges currently faced in Machine Unlearning.


In this competition, we propose a more controllable forgetting loss, EUL, which we integrate with standard Supervised Fine-Tuning (SFT) into a multi-task learning framework. This approach ensures precise unlearning while maximizing the retention of general capabilities. Additionally, we incorporate various data processing and augmentation strategies to further enhance controllable unlearning.

Our experiments demonstrate the effectiveness of EUL and underscored the critical role of standard SFT in training on retained data. Striking the right balance between forgetting and retention is essential for successful unlearning. Furthermore, we introduce an effective data augmentation solution to address the long-tail distribution in text length. Ultimately, our system ranks 5th on the official leaderboard.




\bibliography{custom}

\begin{thebibliography}{14}
\providecommand{\natexlab}[1]{#1}

\bibitem[{Choi et~al.(2024)Choi, Rim, Lee, and Choo}]{choi2024snap}
Minseok Choi, Daniel Rim, Dohyun Lee, and Jaegul Choo. 2024.
\newblock Snap: Unlearning selective knowledge in large language models with negative instructions.
\newblock \emph{arXiv preprint arXiv:2406.12329}.

\bibitem[{Eldan and Russinovich(2023)}]{eldan2023s}
Ronen Eldan and Mark Russinovich. 2023.
\newblock Who's harry potter? approximate unlearning in llms.
\newblock \emph{arXiv preprint arXiv:2310.02238}.

\bibitem[{Groeneveld et~al.(2024)Groeneveld, Beltagy, Walsh, Bhagia, Kinney, Tafjord, Jha, Ivison, Magnusson, Wang et~al.}]{groeneveld2024olmo}
Dirk Groeneveld, Iz~Beltagy, Pete Walsh, Akshita Bhagia, Rodney Kinney, Oyvind Tafjord, Ananya~Harsh Jha, Hamish Ivison, Ian Magnusson, Yizhong Wang, et~al. 2024.
\newblock Olmo: Accelerating the science of language models.
\newblock \emph{arXiv preprint arXiv:2402.00838}.

\bibitem[{Hu et~al.(2022)Hu, Shen, Wallis, Allen-Zhu, Li, Wang, Wang, Chen et~al.}]{hu2022lora}
Edward~J Hu, Yelong Shen, Phillip Wallis, Zeyuan Allen-Zhu, Yuanzhi Li, Shean Wang, Lu~Wang, Weizhu Chen, et~al. 2022.
\newblock Lora: Low-rank adaptation of large language models.
\newblock \emph{ICLR}, 1(2):3.

\bibitem[{Jang et~al.(2022)Jang, Yoon, Yang, Cha, Lee, Logeswaran, and Seo}]{jang2022knowledge}
Joel Jang, Dongkeun Yoon, Sohee Yang, Sungmin Cha, Moontae Lee, Lajanugen Logeswaran, and Minjoon Seo. 2022.
\newblock Knowledge unlearning for mitigating privacy risks in language models.
\newblock \emph{arXiv preprint arXiv:2210.01504}.

\bibitem[{Li et~al.(2025)Li, Zhou, Gao, Chen, Zhang, Kuang, and Fu}]{li2025machine}
Na~Li, Chunyi Zhou, Yansong Gao, Hui Chen, Zhi Zhang, Boyu Kuang, and Anmin Fu. 2025.
\newblock Machine unlearning: Taxonomy, metrics, applications, challenges, and prospects.
\newblock \emph{IEEE Transactions on Neural Networks and Learning Systems}.

\bibitem[{Maini et~al.(2024)Maini, Feng, Schwarzschild, Lipton, and Kolter}]{tofu2024}
Pratyush Maini, Zhili Feng, Avi Schwarzschild, Zachary~C. Lipton, and J.~Zico Kolter. 2024.
\newblock Tofu: A task of fictitious unlearning for llms.

\bibitem[{Mekala et~al.(2024)Mekala, Dorna, Dubey, Lalwani, Koleczek, Rungta, Hasan, and Lobo}]{mekala2024alternate}
Anmol Mekala, Vineeth Dorna, Shreya Dubey, Abhishek Lalwani, David Koleczek, Mukund Rungta, Sadid Hasan, and Elita Lobo. 2024.
\newblock Alternate preference optimization for unlearning factual knowledge in large language models.
\newblock \emph{arXiv preprint arXiv:2409.13474}.

\bibitem[{Qu et~al.(2023)Qu, Yuan, Ding, Ni, Rakotoarivelo, and Smith}]{qu2023learn}
Youyang Qu, Xin Yuan, Ming Ding, Wei Ni, Thierry Rakotoarivelo, and David Smith. 2023.
\newblock Learn to unlearn: A survey on machine unlearning.
\newblock \emph{arXiv preprint arXiv:2305.07512}.

\bibitem[{Ramakrishna et~al.(2025{\natexlab{a}})Ramakrishna, Wan, Jin, Chang, Bu, Vinzamuri, Cevher, Hong, and Gupta}]{ramakrishna2025lumellmunlearningmultitask}
Anil Ramakrishna, Yixin Wan, Xiaomeng Jin, Kai-Wei Chang, Zhiqi Bu, Bhanukiran Vinzamuri, Volkan Cevher, Mingyi Hong, and Rahul Gupta. 2025{\natexlab{a}}.
\newblock Lume: Llm unlearning with multitask evaluations.
\newblock \emph{arXiv preprint arXiv:2502.15097}.

\bibitem[{Ramakrishna et~al.(2025{\natexlab{b}})Ramakrishna, Wan, Jin, Chang, Bu, Vinzamuri, Cevher, Hong, and Gupta}]{ramakrishna2025semeval2025task4unlearning}
Anil Ramakrishna, Yixin Wan, Xiaomeng Jin, Kai-Wei Chang, Zhiqi Bu, Bhanukiran Vinzamuri, Volkan Cevher, Mingyi Hong, and Rahul Gupta. 2025{\natexlab{b}}.
\newblock Semeval-2025 task 4: Unlearning sensitive content from large language models.
\newblock \emph{arXiv preprint arXiv:2504.02883}.

\bibitem[{Shi et~al.(2024)Shi, Tan, Qiu, Qu, Nie, Cheng, Chu, Yinghui, and Qi}]{shi2024ulmr}
Shaojie Shi, Xiaoyu Tan, Xihe Qiu, Chao Qu, Kexin Nie, Yuan Cheng, Wei Chu, Xu~Yinghui, and Yuan Qi. 2024.
\newblock Ulmr: Unlearning large language models via negative response and model parameter average.
\newblock In \emph{Proceedings of the 2024 Conference on Empirical Methods in Natural Language Processing: Industry Track}, pages 755--762.

\bibitem[{Veldanda et~al.(2024)Veldanda, Zhang, Das, Chakraborty, Rawls, Sahu, and Naphade}]{veldanda2024llm}
Akshaj~Kumar Veldanda, Shi-Xiong Zhang, Anirban Das, Supriyo Chakraborty, Stephen Rawls, Sambit Sahu, and Milind Naphade. 2024.
\newblock Llm surgery: Efficient knowledge unlearning and editing in large language models.
\newblock \emph{arXiv preprint arXiv:2409.13054}.

\bibitem[{Wang et~al.(2024)Wang, Tian, Zhang, and Yu}]{wang2024machine}
Weiqi Wang, Zhiyi Tian, Chenhan Zhang, and Shui Yu. 2024.
\newblock Machine unlearning: A comprehensive survey.
\newblock \emph{arXiv preprint arXiv:2405.07406}.

\end{thebibliography}

\appendix
\onecolumn
\section{Appendix: Ablation Study}\nopagebreak[4]
\nopagebreak
\label{sec:appendixA}

\begin{longtable}[h]{|r|r|l|l|l|l|l|r|r|l|r|}
\hline
\multicolumn{1}{|c|}{\textbf{EP}} & \multicolumn{1}{c|}{\textbf{LR}} & \multicolumn{1}{c|}{\textbf{RD}} & \multicolumn{1}{c|}{\textbf{NR}} & \multicolumn{1}{c|}{\textbf{DA}} & \multicolumn{1}{c|}{\textbf{EUL}} & \multicolumn{1}{c|}{\textbf{EUL$^2$ }} & \multicolumn{1}{c|}{\textbf{MIA}} & \multicolumn{1}{c|}{\textbf{TAS}} & \multicolumn{1}{c|}{\textbf{MMLU}} & \multicolumn{1}{c|}{\textbf{Final}} \\ 
\hline
\endfirsthead

\hline
\multicolumn{1}{|c|}{\textbf{EP}} & \multicolumn{1}{c|}{\textbf{LR}} & \multicolumn{1}{c|}{\textbf{RD}} & \multicolumn{1}{c|}{\textbf{NR}} & \multicolumn{1}{c|}{\textbf{DA}} & \multicolumn{1}{c|}{\textbf{EUL}} & \multicolumn{1}{c|}{\textbf{EUL$^2$ }} & \multicolumn{1}{c|}{\textbf{MIA}} & \multicolumn{1}{c|}{\textbf{TAS}} & \multicolumn{1}{c|}{\textbf{MMLU}} & \multicolumn{1}{c|}{\textbf{Final}} \\
\hline
\endhead

\hline
\endfoot

\hline
\endlastfoot
3 & 1.00E-04 & O & O & O & O & × & 0.016 & 0.207 & 0.269 & 0.164 \\ \hline
3 & 1.00E-04 & O & O & O & × & O & 0.014 & 0.095 & 0.272 & 0.127 \\ \hline
3 & 1.00E-04 & O & O & O & × & × & 0.000 & 0.127 & 0.288 & 0.138 \\ \hline
3 & 1.00E-04 & O & O & × & O & × & 0.000 & 0.096 & 0.289 & 0.128 \\ \hline
3 & 1.00E-04 & O & O & × & × & O & 0.000 & 0.092 & 0.279 & 0.124 \\ \hline
3 & 1.00E-04 & O & O & × & × & × & 0.000 & 0.094 & 0.281 & 0.125 \\ \hline
3 & 1.00E-04 & O & × & O & O & × & 0.135 & 0.245 & 0.272 & 0.217 \\ \hline
3 & 1.00E-04 & O & × & O & × & O & 0.051 & 0.109 & 0.274 & 0.145 \\ \hline
3 & 1.00E-04 & O & × & × & O & × & 0.000 & 0.092 & 0.275 & 0.122 \\ \hline
3 & 1.00E-04 & O & × & × & × & O & 0.000 & 0.091 & 0.273 & 0.121 \\ \hline
3 & 1.00E-04 & × & O & O & × & × & 0.000 & 0.091 & 0.274 & 0.122 \\ \hline
3 & 1.00E-04 & × & O & × & O & × & 0.024 & 0.099 & 0.273 & 0.132 \\ \hline
3 & 1.00E-04 & × & O & × & × & × & 0.000 & 0.092 & 0.278 & 0.123 \\ \hline
3 & 1.00E-04 & × & × & O & O & × & 0.996 & 0.424 & 0.229 & 0.550 \\ \hline
3 & 1.00E-04 & × & × & O & × & O & 0.000 & 0.092 & 0.281 & 0.124 \\ \hline
3 & 1.00E-04 & × & × & × & O & × & 0.982 & 0.404 & 0.229 & 0.539 \\ \hline
3 & 1.00E-04 & × & × & × & × & O & 0.940 & 0.395 & 0.246 & 0.527 \\ \hline
3 & 1.00E-05 & O & O & O & O & × & 0.000 & 0.092 & 0.274 & 0.122 \\ \hline
3 & 1.00E-05 & O & O & O & × & O & 0.000 & 0.092 & 0.277 & 0.123 \\ \hline
3 & 1.00E-05 & O & O & O & × & × & 0.000 & 0.092 & 0.274 & 0.122 \\ \hline
3 & 1.00E-05 & O & O & × & O & × & 0.000 & 0.091 & 0.273 & 0.121 \\ \hline
3 & 1.00E-05 & O & O & × & × & O & 0.000 & 0.092 & 0.275 & 0.122 \\ \hline
3 & 1.00E-05 & O & O & × & × & × & 0.000 & 0.091 & 0.272 & 0.121 \\ \hline
3 & 1.00E-05 & O & × & O & O & × & 0.000 & 0.092 & 0.280 & 0.124 \\ \hline
3 & 1.00E-05 & O & × & O & × & O & 0.000 & 0.112 & 0.277 & 0.130 \\ \hline
3 & 1.00E-05 & O & × & × & O & × & 0.000 & 0.094 & 0.281 & 0.125 \\ \hline
3 & 1.00E-05 & O & × & × & × & O & 0.000 & 0.093 & 0.278 & 0.124 \\ \hline
3 & 1.00E-05 & × & O & O & × & × & 0.000 & 0.163 & 0.270 & 0.144 \\ \hline
3 & 1.00E-05 & × & O & × & O & × & 0.000 & 0.091 & 0.273 & 0.121 \\ \hline
3 & 1.00E-05 & × & O & × & × & × & 0.000 & 0.092 & 0.275 & 0.122 \\ \hline
3 & 1.00E-05 & × & × & O & O & × & 0.300 & 0.192 & 0.265 & 0.252 \\ \hline
3 & 1.00E-05 & × & × & O & × & O & 0.000 & 0.092 & 0.280 & 0.124 \\ \hline
3 & 1.00E-05 & × & × & × & O & × & 0.000 & 0.093 & 0.279 & 0.124 \\ \hline
3 & 1.00E-05 & × & × & × & × & O & 0.000 & 0.093 & 0.278 & 0.124 \\ \hline
3 & 1.00E-06 & O & O & O & O & × & 0.000 & 0.092 & 0.277 & 0.123 \\ \hline
3 & 1.00E-06 & O & O & O & × & O & 0.000 & 0.092 & 0.276 & 0.123 \\ \hline
3 & 1.00E-06 & O & O & O & × & × & 0.000 & 0.092 & 0.274 & 0.122 \\ \hline
3 & 1.00E-06 & O & O & × & O & × & 0.000 & 0.091 & 0.274 & 0.122 \\ \hline
3 & 1.00E-06 & O & O & × & × & O & 0.000 & 0.092 & 0.275 & 0.122 \\ \hline
3 & 1.00E-06 & O & O & × & × & × & 0.000 & 0.091 & 0.274 & 0.122 \\ \hline
3 & 1.00E-06 & O & × & O & O & × & 0.000 & 0.092 & 0.275 & 0.122 \\ \hline
3 & 1.00E-06 & O & × & O & × & O & 0.000 & 0.092 & 0.276 & 0.123 \\ \hline
3 & 1.00E-06 & O & × & × & O & × & 0.000 & 0.092 & 0.275 & 0.122 \\ \hline
3 & 1.00E-06 & O & × & × & × & O & 0.000 & 0.092 & 0.276 & 0.123 \\ \hline
3 & 1.00E-06 & × & O & O & × & × & 0.000 & 0.091 & 0.274 & 0.122 \\ \hline
3 & 1.00E-06 & × & O & × & O & × & 0.000 & 0.092 & 0.276 & 0.123 \\ \hline
3 & 1.00E-06 & × & O & × & × & × & 0.000 & 0.092 & 0.274 & 0.122 \\ \hline
3 & 1.00E-06 & × & × & O & O & × & 0.000 & 0.092 & 0.275 & 0.122 \\ \hline
3 & 1.00E-06 & × & × & O & × & O & 0.000 & 0.092 & 0.275 & 0.122 \\ \hline
3 & 1.00E-06 & × & × & × & O & × & 0.000 & 0.092 & 0.275 & 0.122 \\ \hline
3 & 1.00E-06 & × & × & × & × & O & 0.000 & 0.092 & 0.275 & 0.122 \\ \hline
4 & 1.00E-04 & O & O & O & O & × & 0.028 & 0.224 & 0.267 & 0.173 \\ \hline
4 & 1.00E-04 & O & O & O & × & O & 0.021 & 0.224 & 0.275 & 0.173 \\ \hline
4 & 1.00E-04 & O & O & O & × & × & 0.000 & 0.127 & 0.285 & 0.137 \\ \hline
4 & 1.00E-04 & O & O & × & O & × & 0.000 & 0.096 & 0.289 & 0.128 \\ \hline
4 & 1.00E-04 & O & O & × & × & O & 0.000 & 0.094 & 0.281 & 0.125 \\ \hline
4 & 1.00E-04 & O & O & × & × & × & 0.000 & 0.128 & 0.281 & 0.136 \\ \hline
4 & 1.00E-04 & O & × & O & O & × & 0.215 & 0.278 & 0.270 & 0.254 \\ \hline
4 & 1.00E-04 & O & × & O & × & O & 0.219 & 0.165 & 0.274 & 0.219 \\ \hline
4 & 1.00E-04 & O & × & × & O & × & 0.001 & 0.093 & 0.278 & 0.124 \\ \hline
4 & 1.00E-04 & O & × & × & × & O & 0.000 & 0.142 & 0.272 & 0.138 \\ \hline
4 & 1.00E-04 & × & O & O & × & × & 0.000 & 0.090 & 0.271 & 0.121 \\ \hline
4 & 1.00E-04 & × & O & × & O & × & 0.048 & 0.107 & 0.272 & 0.142 \\ \hline
4 & 1.00E-04 & × & O & × & × & × & 0.000 & 0.092 & 0.285 & 0.126 \\ \hline
4 & 1.00E-04 & × & × & O & O & × & 0.993 & 0.423 & 0.229 & 0.548 \\ \hline
4 & 1.00E-04 & × & × & O & × & O & 0.000 & 0.092 & 0.289 & 0.127 \\ \hline
4 & 1.00E-04 & × & × & × & O & × & 0.989 & 0.406 & 0.229 & 0.541 \\ \hline
4 & 1.00E-04 & × & × & × & × & O & 0.945 & 0.397 & 0.246 & 0.529 \\ \hline
4 & 1.00E-05 & O & O & O & O & × & 0.000 & 0.092 & 0.275 & 0.122 \\ \hline
4 & 1.00E-05 & O & O & O & × & O & 0.000 & 0.092 & 0.276 & 0.123 \\ \hline
4 & 1.00E-05 & O & O & O & × & × & 0.000 & 0.092 & 0.275 & 0.122 \\ \hline
4 & 1.00E-05 & O & O & × & O & × & 0.000 & 0.113 & 0.274 & 0.129 \\ \hline
4 & 1.00E-05 & O & O & × & × & O & 0.000 & 0.092 & 0.276 & 0.123 \\ \hline
4 & 1.00E-05 & O & O & × & × & × & 0.000 & 0.091 & 0.274 & 0.122 \\ \hline
4 & 1.00E-05 & O & × & O & O & × & 0.000 & 0.112 & 0.280 & 0.131 \\ \hline
4 & 1.00E-05 & O & × & O & × & O & 0.000 & 0.122 & 0.278 & 0.133 \\ \hline
4 & 1.00E-05 & O & × & × & O & × & 0.000 & 0.093 & 0.280 & 0.124 \\ \hline
4 & 1.00E-05 & O & × & × & × & O & 0.000 & 0.092 & 0.277 & 0.123 \\ \hline
4 & 1.00E-05 & × & O & O & × & × & 0.000 & 0.147 & 0.270 & 0.139 \\ \hline
4 & 1.00E-05 & × & O & × & O & × & 0.000 & 0.177 & 0.274 & 0.150 \\ \hline
4 & 1.00E-05 & × & O & × & × & × & 0.000 & 0.092 & 0.275 & 0.122 \\ \hline
4 & 1.00E-05 & × & × & O & O & × & 0.469 & 0.248 & 0.258 & 0.325 \\ \hline
4 & 1.00E-05 & × & × & O & × & O & 0.000 & 0.114 & 0.281 & 0.132 \\ \hline
4 & 1.00E-05 & × & × & × & O & × & 0.002 & 0.196 & 0.280 & 0.160 \\ \hline
4 & 1.00E-05 & × & × & × & × & O & 0.000 & 0.173 & 0.280 & 0.151 \\ \hline
4 & 1.00E-06 & O & O & O & O & × & 0.000 & 0.092 & 0.276 & 0.123 \\ \hline
4 & 1.00E-06 & O & O & O & × & O & 0.000 & 0.092 & 0.278 & 0.123 \\ \hline
4 & 1.00E-06 & O & O & O & × & × & 0.000 & 0.092 & 0.274 & 0.122 \\ \hline
4 & 1.00E-06 & O & O & × & O & × & 0.000 & 0.092 & 0.275 & 0.122 \\ \hline
4 & 1.00E-06 & O & O & × & × & O & 0.000 & 0.092 & 0.275 & 0.122 \\ \hline
4 & 1.00E-06 & O & O & × & × & × & 0.000 & 0.091 & 0.274 & 0.122 \\ \hline
4 & 1.00E-06 & O & × & O & O & × & 0.000 & 0.092 & 0.276 & 0.122 \\ \hline
4 & 1.00E-06 & O & × & O & × & O & 0.000 & 0.092 & 0.277 & 0.123 \\ \hline
4 & 1.00E-06 & O & × & × & O & × & 0.000 & 0.092 & 0.275 & 0.122 \\ \hline
4 & 1.00E-06 & O & × & × & × & O & 0.000 & 0.092 & 0.277 & 0.123 \\ \hline
4 & 1.00E-06 & × & O & O & × & × & 0.000 & 0.091 & 0.273 & 0.121 \\ \hline
4 & 1.00E-06 & × & O & × & O & × & 0.000 & 0.091 & 0.274 & 0.122 \\ \hline
4 & 1.00E-06 & × & O & × & × & × & 0.000 & 0.092 & 0.274 & 0.122 \\ \hline
4 & 1.00E-06 & × & × & O & O & × & 0.000 & 0.092 & 0.276 & 0.123 \\ \hline
4 & 1.00E-06 & × & × & O & × & O & 0.000 & 0.092 & 0.276 & 0.122 \\ \hline
4 & 1.00E-06 & × & × & × & O & × & 0.000 & 0.092 & 0.276 & 0.123 \\ \hline
4 & 1.00E-06 & × & × & × & × & O & 0.000 & 0.092 & 0.275 & 0.122 \\ \hline
5 & 1.00E-04 & O & O & O & O & × & 0.036 & 0.223 & 0.279 & 0.179 \\ \hline
5 & 1.00E-04 & O & O & O & × & O & 0.022 & 0.226 & 0.279 & 0.175 \\ \hline
5 & 1.00E-04 & O & O & O & × & × & 0.000 & 0.125 & 0.280 & 0.135 \\ \hline
5 & 1.00E-04 & O & O & × & O & × & 0.000 & 0.095 & 0.286 & 0.127 \\ \hline
5 & 1.00E-04 & O & O & × & × & O & 0.000 & 0.096 & 0.287 & 0.127 \\ \hline
5 & 1.00E-04 & O & O & × & × & × & 0.000 & 0.123 & 0.279 & 0.134 \\ \hline
5 & 1.00E-04 & O & × & O & O & × & 0.593 & 0.395 & 0.275 & 0.421 \\ \hline
5 & 1.00E-04 & O & × & O & × & O & 0.387 & 0.221 & 0.276 & 0.295 \\ \hline
5 & 1.00E-04 & O & × & × & O & × & 0.005 & 0.194 & 0.275 & 0.158 \\ \hline
5 & 1.00E-04 & O & × & × & × & O & 0.001 & 0.147 & 0.271 & 0.140 \\ \hline
5 & 1.00E-04 & × & O & O & × & × & 0.000 & 0.093 & 0.278 & 0.124 \\ \hline
5 & 1.00E-04 & × & O & × & O & × & 0.010 & 0.095 & 0.276 & 0.127 \\ \hline
5 & 1.00E-04 & × & O & × & × & × & 0.000 & 0.092 & 0.281 & 0.124 \\ \hline
5 & 1.00E-04 & × & × & O & O & × & 0.989 & 0.421 & 0.229 & 0.547 \\ \hline
5 & 1.00E-04 & × & × & O & × & O & 0.000 & 0.092 & 0.291 & 0.128 \\ \hline
5 & 1.00E-04 & × & × & × & O & × & 0.991 & 0.407 & 0.229 & 0.543 \\ \hline
5 & 1.00E-04 & × & × & × & × & O & 0.947 & 0.396 & 0.240 & 0.528 \\ \hline
5 & 1.00E-05 & O & O & O & O & × & 0.000 & 0.092 & 0.276 & 0.123 \\ \hline
5 & 1.00E-05 & O & O & O & × & O & 0.000 & 0.092 & 0.276 & 0.123 \\ \hline
5 & 1.00E-05 & O & O & O & × & × & 0.000 & 0.092 & 0.276 & 0.123 \\ \hline
5 & 1.00E-05 & O & O & × & O & × & 0.000 & 0.180 & 0.276 & 0.152 \\ \hline
5 & 1.00E-05 & O & O & × & × & O & 0.000 & 0.092 & 0.275 & 0.122 \\ \hline
5 & 1.00E-05 & O & O & × & × & × & 0.000 & 0.092 & 0.275 & 0.122 \\ \hline
5 & 1.00E-05 & O & × & O & O & × & 0.001 & 0.112 & 0.279 & 0.131 \\ \hline
5 & 1.00E-05 & O & × & O & × & O & 0.000 & 0.112 & 0.277 & 0.130 \\ \hline
5 & 1.00E-05 & O & × & × & O & × & 0.000 & 0.093 & 0.280 & 0.125 \\ \hline
5 & 1.00E-05 & O & × & × & × & O & 0.000 & 0.114 & 0.276 & 0.130 \\ \hline
5 & 1.00E-05 & × & O & O & × & × & 0.000 & 0.090 & 0.270 & 0.120 \\ \hline
5 & 1.00E-05 & × & O & × & O & × & 0.000 & 0.201 & 0.273 & 0.158 \\ \hline
5 & 1.00E-05 & × & O & × & × & × & 0.000 & 0.147 & 0.272 & 0.140 \\ \hline
5 & 1.00E-05 & × & × & O & O & × & 0.569 & 0.281 & 0.257 & 0.369 \\ \hline
5 & 1.00E-05 & × & × & O & × & O & 0.000 & 0.125 & 0.280 & 0.135 \\ \hline
5 & 1.00E-05 & × & × & × & O & × & 0.141 & 0.138 & 0.273 & 0.184 \\ \hline
5 & 1.00E-05 & × & × & × & × & O & 0.002 & 0.194 & 0.279 & 0.158 \\ \hline
5 & 1.00E-06 & O & O & O & O & × & 0.000 & 0.092 & 0.276 & 0.123 \\ \hline
5 & 1.00E-06 & O & O & O & × & O & 0.000 & 0.092 & 0.277 & 0.123 \\ \hline
5 & 1.00E-06 & O & O & O & × & × & 0.000 & 0.092 & 0.273 & 0.122 \\ \hline
5 & 1.00E-06 & O & O & × & O & × & 0.000 & 0.091 & 0.274 & 0.122 \\ \hline
5 & 1.00E-06 & O & O & × & × & O & 0.000 & 0.092 & 0.275 & 0.122 \\ \hline
5 & 1.00E-06 & O & O & × & × & × & 0.000 & 0.091 & 0.274 & 0.122 \\ \hline
5 & 1.00E-06 & O & × & O & O & × & 0.000 & 0.092 & 0.277 & 0.123 \\ \hline
5 & 1.00E-06 & O & × & O & × & O & 0.000 & 0.092 & 0.278 & 0.123 \\ \hline
5 & 1.00E-06 & O & × & × & O & × & 0.000 & 0.092 & 0.275 & 0.122 \\ \hline
5 & 1.00E-06 & O & × & × & × & O & 0.000 & 0.091 & 0.274 & 0.122 \\ \hline
5 & 1.00E-06 & × & O & O & × & × & 0.000 & 0.091 & 0.273 & 0.121 \\ \hline
5 & 1.00E-06 & × & O & × & O & × & 0.000 & 0.091 & 0.273 & 0.122 \\ \hline
5 & 1.00E-06 & × & O & × & × & × & 0.000 & 0.092 & 0.274 & 0.122 \\ \hline
5 & 1.00E-06 & × & × & O & O & × & 0.000 & 0.092 & 0.277 & 0.123 \\ \hline
5 & 1.00E-06 & × & × & O & × & O & 0.000 & 0.092 & 0.276 & 0.122 \\ \hline
5 & 1.00E-06 & × & × & × & O & × & 0.000 & 0.092 & 0.275 & 0.122 \\ \hline
5 & 1.00E-06 & × & × & × & × & O & 0.000 & 0.092 & 0.275 & 0.122 \\ \hline

\caption{Ablation Study Results in OLMo-1B}
\label{sec:tab_app_ablation}
\end{longtable}

\begin{longtable}[h]{|r|r|l|l|l|l|l|r|r|l|r|}
\hline
\multicolumn{1}{|c|}{\textbf{EP}} & \multicolumn{1}{c|}{\textbf{LR}} & \multicolumn{1}{c|}{\textbf{RD}} & \multicolumn{1}{c|}{\textbf{NR}} & \multicolumn{1}{c|}{\textbf{DA}} & \multicolumn{1}{c|}{\textbf{EUL}} & \multicolumn{1}{c|}{\textbf{EUL$^2$ }} & \multicolumn{1}{c|}{\textbf{MIA}} & \multicolumn{1}{c|}{\textbf{TAS}} & \multicolumn{1}{c|}{\textbf{MMLU}} & \multicolumn{1}{c|}{\textbf{Final}} \\ 
\hline
\endfirsthead

\hline
\multicolumn{1}{|c|}{\textbf{EP}} & \multicolumn{1}{c|}{\textbf{LR}} & \multicolumn{1}{c|}{\textbf{RD}} & \multicolumn{1}{c|}{\textbf{NR}} & \multicolumn{1}{c|}{\textbf{DA}} & \multicolumn{1}{c|}{\textbf{EUL}} & \multicolumn{1}{c|}{\textbf{EUL$^2$ }} & \multicolumn{1}{c|}{\textbf{MIA}} & \multicolumn{1}{c|}{\textbf{TAS}} & \multicolumn{1}{c|}{\textbf{MMLU}} & \multicolumn{1}{c|}{\textbf{Final}} \\
\hline
\endhead

\hline
\endfoot

\hline
\endlastfoot
3 & 1.00E-04 & × & × & × & × & O & 0.992 & 0.407 & 0.229 & 0.543 \\ \hline
3 & 1.00E-04 & × & × & × & O & × & 0.991 & 0.407 & 0.229 & 0.543 \\ \hline
3 & 1.00E-04 & × & × & O & × & O & 0.958 & 0.409 & 0.269 & 0.545 \\ \hline
3 & 1.00E-04 & × & × & O & O & × & 0.959 & 0.409 & 0.269 & 0.546 \\ \hline
3 & 1.00E-04 & × & O & × & × & × & 0.000 & 0.165 & 0.494 & 0.220 \\ \hline
3 & 1.00E-04 & × & O & × & O & × & 0.165 & 0.218 & 0.490 & 0.291 \\ \hline
3 & 1.00E-04 & × & O & O & × & × & 0.000 & 0.165 & 0.496 & 0.220 \\ \hline
3 & 1.00E-04 & O & × & × & × & O & 0.010 & 0.229 & 0.496 & 0.245 \\ \hline
3 & 1.00E-04 & O & × & × & O & × & 0.063 & 0.196 & 0.497 & 0.252 \\ \hline
3 & 1.00E-04 & O & × & O & × & O & 0.420 & 0.327 & 0.496 & 0.414 \\ \hline
3 & 1.00E-04 & O & × & O & O & × & 0.131 & 0.300 & 0.498 & 0.310 \\ \hline
3 & 1.00E-04 & O & O & × & × & × & 0.000 & 0.167 & 0.500 & 0.222 \\ \hline
3 & 1.00E-04 & O & O & × & × & O & 0.000 & 0.165 & 0.496 & 0.221 \\ \hline
3 & 1.00E-04 & O & O & × & O & × & 0.000 & 0.161 & 0.484 & 0.215 \\ \hline
3 & 1.00E-04 & O & O & O & × & × & 0.000 & 0.166 & 0.497 & 0.221 \\ \hline
3 & 1.00E-04 & O & O & O & × & O & 0.060 & 0.242 & 0.506 & 0.269 \\ \hline
3 & 1.00E-04 & O & O & O & O & × & 0.056 & 0.192 & 0.499 & 0.249 \\ \hline
3 & 1.00E-05 & × & × & × & × & O & 0.000 & 0.318 & 0.499 & 0.272 \\ \hline
3 & 1.00E-05 & × & × & × & O & × & 0.000 & 0.310 & 0.494 & 0.268 \\ \hline
3 & 1.00E-05 & × & × & O & × & O & 0.839 & 0.361 & 0.244 & 0.481 \\ \hline
3 & 1.00E-05 & × & × & O & O & × & 0.937 & 0.389 & 0.230 & 0.519 \\ \hline
3 & 1.00E-05 & × & O & × & × & × & 0.000 & 0.260 & 0.504 & 0.255 \\ \hline
3 & 1.00E-05 & × & O & × & O & × & 0.000 & 0.230 & 0.498 & 0.243 \\ \hline
3 & 1.00E-05 & × & O & O & × & × & 0.000 & 0.203 & 0.499 & 0.234 \\ \hline
3 & 1.00E-05 & O & × & × & × & O & 0.000 & 0.168 & 0.504 & 0.224 \\ \hline
3 & 1.00E-05 & O & × & × & O & × & 0.000 & 0.168 & 0.503 & 0.223 \\ \hline
3 & 1.00E-05 & O & × & O & × & O & 0.010 & 0.167 & 0.491 & 0.222 \\ \hline
3 & 1.00E-05 & O & × & O & O & × & 0.028 & 0.172 & 0.488 & 0.229 \\ \hline
3 & 1.00E-05 & O & O & × & × & × & 0.000 & 0.168 & 0.504 & 0.224 \\ \hline
3 & 1.00E-05 & O & O & × & × & O & 0.000 & 0.235 & 0.505 & 0.246 \\ \hline
3 & 1.00E-05 & O & O & × & O & × & 0.000 & 0.235 & 0.505 & 0.246 \\ \hline
3 & 1.00E-05 & O & O & O & × & × & 0.000 & 0.167 & 0.501 & 0.223 \\ \hline
3 & 1.00E-05 & O & O & O & × & O & 0.000 & 0.262 & 0.497 & 0.253 \\ \hline
3 & 1.00E-05 & O & O & O & O & × & 0.000 & 0.255 & 0.496 & 0.250 \\ \hline
3 & 1.00E-06 & × & × & × & × & O & 0.000 & 0.170 & 0.509 & 0.226 \\ \hline
3 & 1.00E-06 & × & × & × & O & × & 0.000 & 0.169 & 0.508 & 0.226 \\ \hline
3 & 1.00E-06 & × & × & O & × & O & 0.000 & 0.169 & 0.508 & 0.226 \\ \hline
3 & 1.00E-06 & × & × & O & O & × & 0.000 & 0.169 & 0.508 & 0.226 \\ \hline
3 & 1.00E-06 & × & O & × & × & × & 0.000 & 0.170 & 0.509 & 0.226 \\ \hline
3 & 1.00E-06 & × & O & × & O & × & 0.000 & 0.170 & 0.510 & 0.226 \\ \hline
3 & 1.00E-06 & × & O & O & × & × & 0.000 & 0.170 & 0.510 & 0.227 \\ \hline
3 & 1.00E-06 & O & × & × & × & O & 0.000 & 0.170 & 0.510 & 0.227 \\ \hline
3 & 1.00E-06 & O & × & × & O & × & 0.000 & 0.169 & 0.508 & 0.226 \\ \hline
3 & 1.00E-06 & O & × & O & × & O & 0.000 & 0.170 & 0.509 & 0.226 \\ \hline
3 & 1.00E-06 & O & × & O & O & × & 0.000 & 0.170 & 0.509 & 0.226 \\ \hline
3 & 1.00E-06 & O & O & × & × & × & 0.000 & 0.170 & 0.510 & 0.227 \\ \hline
3 & 1.00E-06 & O & O & × & × & O & 0.000 & 0.170 & 0.510 & 0.227 \\ \hline
3 & 1.00E-06 & O & O & × & O & × & 0.000 & 0.170 & 0.509 & 0.226 \\ \hline
3 & 1.00E-06 & O & O & O & × & × & 0.000 & 0.170 & 0.509 & 0.226 \\ \hline
3 & 1.00E-06 & O & O & O & × & O & 0.000 & 0.170 & 0.510 & 0.227 \\ \hline
3 & 1.00E-06 & O & O & O & O & × & 0.000 & 0.170 & 0.509 & 0.226 \\ \hline
4 & 1.00E-04 & × & × & × & × & O & 0.991 & 0.407 & 0.229 & 0.542 \\ \hline
4 & 1.00E-04 & × & × & × & O & × & 0.988 & 0.406 & 0.229 & 0.541 \\ \hline
4 & 1.00E-04 & × & × & O & × & O & 0.956 & 0.408 & 0.269 & 0.544 \\ \hline
4 & 1.00E-04 & × & × & O & O & × & 0.961 & 0.410 & 0.269 & 0.546 \\ \hline
4 & 1.00E-04 & × & O & × & × & × & 0.000 & 0.164 & 0.492 & 0.219 \\ \hline
4 & 1.00E-04 & × & O & × & O & × & 0.135 & 0.212 & 0.502 & 0.283 \\ \hline
4 & 1.00E-04 & × & O & O & × & × & 0.000 & 0.166 & 0.497 & 0.221 \\ \hline
4 & 1.00E-04 & O & × & × & × & O & 0.116 & 0.316 & 0.492 & 0.308 \\ \hline
4 & 1.00E-04 & O & × & × & O & × & 0.133 & 0.293 & 0.482 & 0.303 \\ \hline
4 & 1.00E-04 & O & × & O & × & O & 0.461 & 0.417 & 0.487 & 0.455 \\ \hline
4 & 1.00E-04 & O & × & O & O & × & 0.712 & 0.462 & 0.476 & 0.550 \\ \hline
4 & 1.00E-04 & O & O & × & × & × & 0.000 & 0.164 & 0.493 & 0.219 \\ \hline
4 & 1.00E-04 & O & O & × & × & O & 0.000 & 0.205 & 0.495 & 0.233 \\ \hline
4 & 1.00E-04 & O & O & × & O & × & 0.000 & 0.223 & 0.496 & 0.239 \\ \hline
4 & 1.00E-04 & O & O & O & × & × & 0.000 & 0.164 & 0.493 & 0.219 \\ \hline
4 & 1.00E-04 & O & O & O & × & O & 0.082 & 0.194 & 0.501 & 0.259 \\ \hline
4 & 1.00E-04 & O & O & O & O & × & 0.048 & 0.237 & 0.501 & 0.262 \\ \hline
4 & 1.00E-05 & × & × & × & × & O & 0.719 & 0.378 & 0.414 & 0.504 \\ \hline
4 & 1.00E-05 & × & × & × & O & × & 0.760 & 0.371 & 0.351 & 0.494 \\ \hline
4 & 1.00E-05 & × & × & O & × & O & 0.860 & 0.366 & 0.239 & 0.488 \\ \hline
4 & 1.00E-05 & × & × & O & O & × & 0.965 & 0.398 & 0.229 & 0.531 \\ \hline
4 & 1.00E-05 & × & O & × & × & × & 0.000 & 0.206 & 0.501 & 0.236 \\ \hline
4 & 1.00E-05 & × & O & × & O & × & 0.005 & 0.165 & 0.491 & 0.221 \\ \hline
4 & 1.00E-05 & × & O & O & × & × & 0.000 & 0.190 & 0.497 & 0.229 \\ \hline
4 & 1.00E-05 & O & × & × & × & O & 0.000 & 0.166 & 0.499 & 0.222 \\ \hline
4 & 1.00E-05 & O & × & × & O & × & 0.000 & 0.166 & 0.499 & 0.222 \\ \hline
4 & 1.00E-05 & O & × & O & × & O & 0.082 & 0.192 & 0.492 & 0.255 \\ \hline
4 & 1.00E-05 & O & × & O & O & × & 0.237 & 0.241 & 0.487 & 0.322 \\ \hline
4 & 1.00E-05 & O & O & × & × & × & 0.000 & 0.167 & 0.502 & 0.223 \\ \hline
4 & 1.00E-05 & O & O & × & × & O & 0.000 & 0.209 & 0.503 & 0.237 \\ \hline
4 & 1.00E-05 & O & O & × & O & × & 0.000 & 0.208 & 0.504 & 0.237 \\ \hline
4 & 1.00E-05 & O & O & O & × & × & 0.000 & 0.166 & 0.499 & 0.222 \\ \hline
4 & 1.00E-05 & O & O & O & × & O & 0.000 & 0.261 & 0.493 & 0.252 \\ \hline
4 & 1.00E-05 & O & O & O & O & × & 0.000 & 0.290 & 0.493 & 0.261 \\ \hline
4 & 1.00E-06 & × & × & × & × & O & 0.000 & 0.170 & 0.509 & 0.226 \\ \hline
4 & 1.00E-06 & × & × & × & O & × & 0.000 & 0.169 & 0.508 & 0.226 \\ \hline
4 & 1.00E-06 & × & × & O & × & O & 0.000 & 0.169 & 0.507 & 0.225 \\ \hline
4 & 1.00E-06 & × & × & O & O & × & 0.000 & 0.169 & 0.507 & 0.225 \\ \hline
4 & 1.00E-06 & × & O & × & × & × & 0.000 & 0.170 & 0.509 & 0.226 \\ \hline
4 & 1.00E-06 & × & O & × & O & × & 0.000 & 0.169 & 0.508 & 0.226 \\ \hline
4 & 1.00E-06 & × & O & O & × & × & 0.000 & 0.170 & 0.509 & 0.226 \\ \hline
4 & 1.00E-06 & O & × & × & × & O & 0.000 & 0.170 & 0.509 & 0.226 \\ \hline
4 & 1.00E-06 & O & × & × & O & × & 0.000 & 0.170 & 0.509 & 0.226 \\ \hline
4 & 1.00E-06 & O & × & O & × & O & 0.000 & 0.169 & 0.507 & 0.225 \\ \hline
4 & 1.00E-06 & O & × & O & O & × & 0.000 & 0.169 & 0.508 & 0.226 \\ \hline
4 & 1.00E-06 & O & O & × & × & × & 0.000 & 0.170 & 0.509 & 0.226 \\ \hline
4 & 1.00E-06 & O & O & × & × & O & 0.000 & 0.169 & 0.508 & 0.226 \\ \hline
4 & 1.00E-06 & O & O & × & O & × & 0.000 & 0.170 & 0.509 & 0.226 \\ \hline
4 & 1.00E-06 & O & O & O & × & × & 0.000 & 0.170 & 0.510 & 0.227 \\ \hline
4 & 1.00E-06 & O & O & O & × & O & 0.000 & 0.170 & 0.509 & 0.226 \\ \hline
4 & 1.00E-06 & O & O & O & O & × & 0.000 & 0.170 & 0.509 & 0.226 \\ \hline
5 & 1.00E-04 & × & × & × & × & O & 0.993 & 0.407 & 0.229 & 0.543 \\ \hline
5 & 1.00E-04 & × & × & × & O & × & 0.984 & 0.405 & 0.229 & 0.539 \\ \hline
5 & 1.00E-04 & × & × & O & × & O & 0.955 & 0.408 & 0.269 & 0.544 \\ \hline
5 & 1.00E-04 & × & × & O & O & × & 0.959 & 0.409 & 0.269 & 0.546 \\ \hline
5 & 1.00E-04 & × & O & × & × & × & 0.000 & 0.163 & 0.488 & 0.217 \\ \hline
5 & 1.00E-04 & × & O & × & O & × & 0.726 & 0.319 & 0.230 & 0.425 \\ \hline
5 & 1.00E-04 & × & O & O & × & × & 0.000 & 0.163 & 0.488 & 0.217 \\ \hline
5 & 1.00E-04 & O & × & × & × & O & 0.285 & 0.361 & 0.482 & 0.376 \\ \hline
5 & 1.00E-04 & O & × & × & O & × & 0.318 & 0.282 & 0.495 & 0.365 \\ \hline
5 & 1.00E-04 & O & × & O & × & O & 0.559 & 0.431 & 0.500 & 0.497 \\ \hline
5 & 1.00E-04 & O & × & O & O & × & 0.747 & 0.529 & 0.466 & 0.581 \\ \hline
5 & 1.00E-04 & O & O & × & × & × & 0.000 & 0.164 & 0.492 & 0.219 \\ \hline
5 & 1.00E-04 & O & O & × & × & O & 0.000 & 0.229 & 0.500 & 0.243 \\ \hline
5 & 1.00E-04 & O & O & × & O & × & 0.000 & 0.186 & 0.492 & 0.226 \\ \hline
5 & 1.00E-04 & O & O & O & × & × & 0.000 & 0.175 & 0.494 & 0.223 \\ \hline
5 & 1.00E-04 & O & O & O & × & O & 0.066 & 0.192 & 0.489 & 0.249 \\ \hline
5 & 1.00E-04 & O & O & O & O & × & 0.038 & 0.209 & 0.498 & 0.248 \\ \hline
5 & 1.00E-05 & × & × & × & × & O & 0.791 & 0.370 & 0.318 & 0.493 \\ \hline
5 & 1.00E-05 & × & × & × & O & × & 0.915 & 0.386 & 0.244 & 0.515 \\ \hline
5 & 1.00E-05 & × & × & O & × & O & 0.879 & 0.372 & 0.237 & 0.496 \\ \hline
5 & 1.00E-05 & × & × & O & O & × & 0.972 & 0.401 & 0.229 & 0.534 \\ \hline
5 & 1.00E-05 & × & O & × & × & × & 0.000 & 0.166 & 0.498 & 0.222 \\ \hline
5 & 1.00E-05 & × & O & × & O & × & 0.022 & 0.168 & 0.483 & 0.225 \\ \hline
5 & 1.00E-05 & × & O & O & × & × & 0.000 & 0.188 & 0.492 & 0.227 \\ \hline
5 & 1.00E-05 & O & × & × & × & O & 0.000 & 0.165 & 0.494 & 0.219 \\ \hline
5 & 1.00E-05 & O & × & × & O & × & 0.000 & 0.165 & 0.494 & 0.219 \\ \hline
5 & 1.00E-05 & O & × & O & × & O & 0.240 & 0.315 & 0.492 & 0.349 \\ \hline
5 & 1.00E-05 & O & × & O & O & × & 0.594 & 0.360 & 0.487 & 0.480 \\ \hline
5 & 1.00E-05 & O & O & × & × & × & 0.000 & 0.167 & 0.502 & 0.223 \\ \hline
5 & 1.00E-05 & O & O & × & × & O & 0.000 & 0.207 & 0.502 & 0.236 \\ \hline
5 & 1.00E-05 & O & O & × & O & × & 0.000 & 0.209 & 0.501 & 0.237 \\ \hline
5 & 1.00E-05 & O & O & O & × & × & 0.000 & 0.187 & 0.498 & 0.228 \\ \hline
5 & 1.00E-05 & O & O & O & × & O & 0.000 & 0.240 & 0.497 & 0.246 \\ \hline
5 & 1.00E-05 & O & O & O & O & × & 0.000 & 0.290 & 0.494 & 0.262 \\ \hline
5 & 1.00E-06 & × & × & × & × & O & 0.000 & 0.170 & 0.510 & 0.227 \\ \hline
5 & 1.00E-06 & × & × & × & O & × & 0.000 & 0.169 & 0.508 & 0.226 \\ \hline
5 & 1.00E-06 & × & × & O & × & O & 0.000 & 0.169 & 0.507 & 0.225 \\ \hline
5 & 1.00E-06 & × & × & O & O & × & 0.000 & 0.169 & 0.507 & 0.226 \\ \hline
5 & 1.00E-06 & × & O & × & × & × & 0.000 & 0.170 & 0.510 & 0.227 \\ \hline
5 & 1.00E-06 & × & O & × & O & × & 0.000 & 0.170 & 0.509 & 0.226 \\ \hline
5 & 1.00E-06 & × & O & O & × & × & 0.000 & 0.170 & 0.509 & 0.226 \\ \hline
5 & 1.00E-06 & O & × & × & × & O & 0.000 & 0.169 & 0.508 & 0.226 \\ \hline
5 & 1.00E-06 & O & × & × & O & × & 0.000 & 0.169 & 0.508 & 0.226 \\ \hline
5 & 1.00E-06 & O & × & O & × & O & 0.000 & 0.169 & 0.507 & 0.225 \\ \hline
5 & 1.00E-06 & O & × & O & O & × & 0.000 & 0.169 & 0.508 & 0.226 \\ \hline
5 & 1.00E-06 & O & O & × & × & × & 0.000 & 0.170 & 0.509 & 0.226 \\ \hline
5 & 1.00E-06 & O & O & × & × & O & 0.000 & 0.170 & 0.509 & 0.226 \\ \hline
5 & 1.00E-06 & O & O & × & O & × & 0.000 & 0.170 & 0.509 & 0.226 \\ \hline
5 & 1.00E-06 & O & O & O & × & × & 0.000 & 0.170 & 0.510 & 0.226 \\ \hline
5 & 1.00E-06 & O & O & O & × & O & 0.000 & 0.170 & 0.509 & 0.226 \\ \hline
5 & 1.00E-06 & O & O & O & O & × & 0.000 & 0.169 & 0.508 & 0.226 \\ \hline

\caption{Ablation Study Results in OLMo-7B}
\label{sec:tab_app_ablation_7B}
\end{longtable}

\end{document}